# *FingerPrint*: A 3-D Printed Soft Monolithic 4-Degree-of-Freedom Fingertip Haptic Device with Embedded Actuation

Zhenishbek Zhakypov and Allison M. Okamura

*Abstract*— Wearable fingertip haptic interfaces provide tactile stimuli on the fingerpads by applying skin pressure, linear and rotational shear, and vibration. Designing and fabricating a compact, multi-degree-of-freedom, and forceful fingertip haptic interface is challenging due to trade-offs among miniaturization, multifunctionality, and manufacturability. Downsizing electromagnetic actuators that produce high torques is infeasible, and integrating multiple actuators, links, joints, and transmission elements increases device size and weight. 3-D printing enables rapid manufacturing of complex devices with minimal assembly in large batches. However, it requires a careful arrangement of material properties, geometry, scale, and printer capabilities. Here we present a fully 3-D printed, soft, monolithic fingertip haptic device based on an origami pattern known as the "waterbomb" base that embeds foldable vacuum actuation and produces 4-DoF of motion on the fingerpad with tunable haptic forces (up to 1.3 N shear and 7 N normal) and torque (up to 25 N-mm). Including the thimble mounting, the compact device is 40 mm long and 20 mm wide. This demonstrates the efficacy of origami design and soft material 3D printing for designing and rapidly fabricating miniature yet complex wearable mechanisms with force output appropriate for haptic interaction.

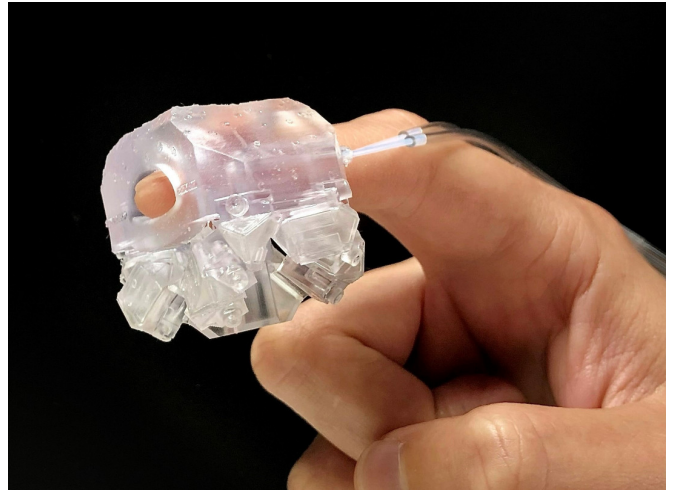

Fig. 1. *FingerPrint* is a novel soft wearable 4-DoF haptic device for the fingertip that is 3-D printed and uses pneumatic actuation to generate pressure, linear and rotational shear, and vibration stimuli on the skin.

## I. INTRODUCTION

Haptic (touch-based) perception is critical to how we perceive physical properties of objects and the environment. Haptic feedback is a key component of the sensorimotor experience, yet in consumer devices, haptic stimuli are typically limited to simple vibrations used as event alerts that call the attention of the user, rather than enable immersion or increase information transfer rate. Thousands of mechanoreceptors within the fingerpads capture a wide range of information by skin deformation, such as shear, pressure, and vibration, to distinguish various object features through exploratory procedures. For example, one can apply pressure to an object with the index finger and thumb to determine its stiffness, lift and turn it in the air to determine its weight and rotational inertia, and slide the fingertips on its surface to feel its friction, texture, and contours [1]. Fingertip haptic devices localize stimulation on the most touch-sensitive and useful area of human skin – the fingerpads – by providing multi-degree-of-freedom (DoF) skin deformation and forces at the mesoscale (millimeter scale) [2]. Wearable fingertip haptic devices can be used to generate realistic force feedback without the need for world-grounded forces [3]. Engineering a versatile, compelling, and safe haptic interface for rendering such complex interactions at the fingertip is a significant design challenge, which we have addressed through *FingerPrint*, a novel 3-D printed wearable fingertip haptic device (Fig. 1).

Conventional robotic systems for mesoscale haptic applications face fundamental challenges. They incorporate motors, transmission elements, rigid links, joints, and other mechanical components to enable multi-DoF motions. So, it requires a multitude of diverse elements integrated into a compact and lightweight design. Downsizing conventional electromagnetic actuators, e.g., direct current (DC) motors, coupled with gear trains to produce forces on the order of Newtons is also infeasible. The classical joints that combine two or more kinematic pairs, such as pin-hole, ball-socket, and slider-slot, further hinder miniaturization, manufacturing, and assembly [4].

Existing designs for multi-DoF fingertip haptic interfaces have integrated standard components, resulting in relatively bulky and complex construction and high cost [5], [6], [7], [8]. Some of these devices considerably reduce the DoFs [9], [10] or rely only on the limited modality of feedback by vibrotactile illusions [11], [12], neglecting the wide variety of possible tactile stimuli. Recent studies on foldable mechanisms using compliant joints and multi-layer composite fabrication techniques present opportunities for multifunctional yet compact fingertip tactile interface designs [13], [14]. However, these prototypes still employ off-the-

*This work is supported in part by the Swiss National Science Foundation (SNSF) Early Postdoc Mobility grant P2ELP2 195132 and US National Science Foundation grants 1830163 and 1812966.

All authors are with the department of Mechanical Engineering, Stanford University, Stanford, CA 94301, USA. zhakypov@stanford.edu, aokamura@stanford.edu

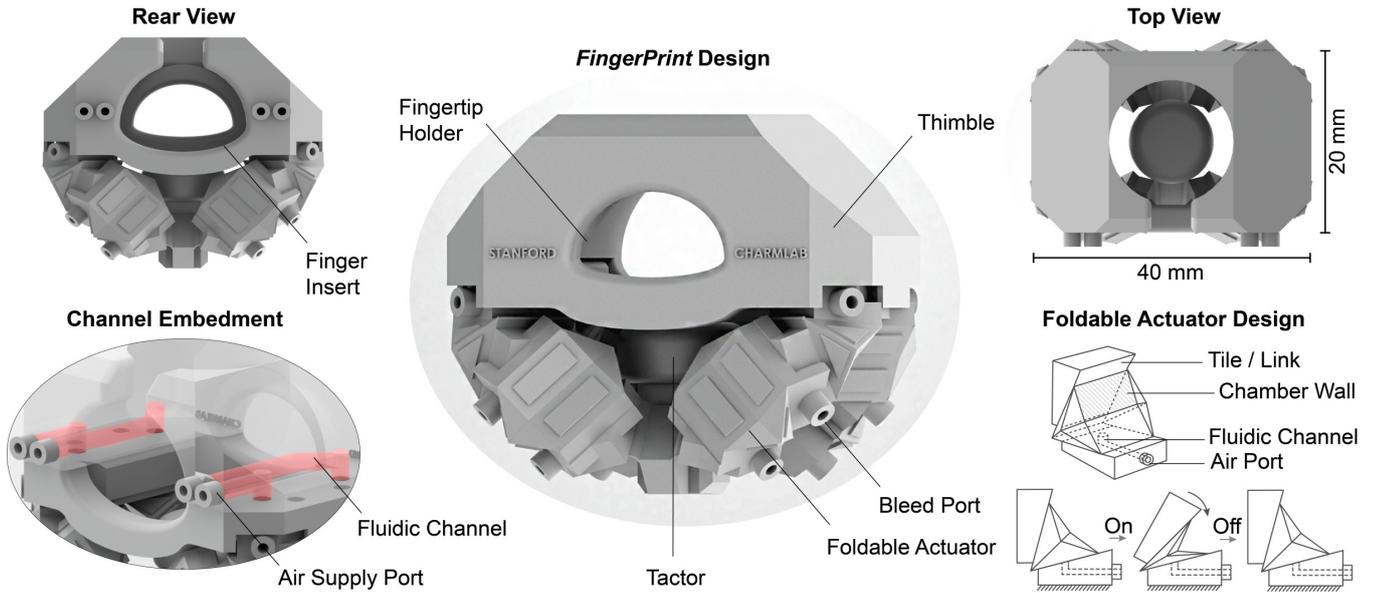

Fig. 2. The monolithic design of the *FingerPrint* haptic device, showing the components including the tactor, foldable actuators, fluidic channels, pneumatic supply ports, and pneumatic bleed ports. The entire device is worn by placing a finger through the insert on the thimble (fingertip holder). The airtight, origami-based foldable actuator collapses and the joint folds upon vacuum pressure (negative) input to the supply port.

shelf electromagnetic or piezoelecric motors that limit miniaturization and require complex manual assembly processes.

Alternatively, three-dimensional (3-D) printing offers high freedom and speed for fabricating complex soft and compliant mechanical structures, actuators, and mechanisms, ideally with the push of a button [15]. Proposed 3-D printed fingertip haptic prototypes provide stimuli by single [16] or distributed [17] one-DoF inflatable actuators for simple interaction. Despite its potential, little research has been conducted into 3-D printing technology for multi-DoF mesoscale haptic interfaces.

Here we present a fully 3-D printed, soft, monolithic 4-DoF fingertip haptic device, called *FingerPrint*, that stimulates linear and rotational shear, pressure, and vibration on the fingerpad. Constructed using an origami "waterbomb" base mechanism and printed from a flexible material, *FingerPrint* embeds eight foldable vacuum-powered pneumatic actuators to achieve three translational ($x$, $y$, $z$) and one rotational (twist) tactile motions and forces of a tactor end-effector on the fingerpad skin (Fig. 2). The tactor produces $\pm 3$ mm and $\pm 1.3$ N in $x$, $\pm 2$ mm and $\pm 1.2$ N in $y$, and $\pm 4$ mm and $\pm 7$ N in $z$ Cartesian coordinates, and $\pm 25$ degrees and $\pm 25$ N-mm torque in rotation (yaw). The soft device gently interfaces with a user's finger via a soft thimble and embeds multiple fluidic channels for vacuum supply. This work advances assembly-free mass fabrication of miniature and multifunctional haptic devices.

The manuscript is structured as follows. In Section II, we describe the design steps for our device. We present the fabrication methodology in Section III. We experimentally characterize the interface for motion, forces, and frequency response in Section IV. We provide concluding remarks and discuss future design improvements in Section V.

## II. FINGERTIP DEVICE DESIGN

### A. Design for 3-D Printing

3-D printing requires meticulous attention to the model design, materials, and post-processing procedures at the outset to achieve accurate and functional prototypes. The choice of printing technology is a crucial factor. Stereolithography (SLA) and Fused Deposition Modeling (FDM) are two broadly available 3-D printing methods that are relatively affordable. SLA employs a traveling laser beam to cure layers of light-reactive resin by photopolymerization, whereas FDM melts and deposits thermoplastic filaments through a heated nozzle. In contrast to FDM, SLA enables high resolution, isotropy, complex geometries, and a great variety of material choices, from rigid plastics to flexible resins. Laser optical spot size produces fine model features and applies minimal force, making it convenient for mesoscale manufacturing. SLA 3D printers, however, cannot inherently print multiple materials at once, so varying mechanical properties can be achieved using variable material thickness. The critical factors for SLA include machine capabilities, utilized material properties, print scale and geometry, model and support placement, arrangement of bleeds for trapped resin, washing and curing conditions.

We used an SLA 3-D printer (Formlabs Form 3) to design and fabricate the mesoscale fingertip haptic device. To embed multifunctionality, we choose a flexible resin (Flexible 80A, Formlabs) due to its high flexibility, durability, and printing accuracy (50-100 $\mu$m). This resin's proven mechanical characteristics permit repeated bending, flexing, and compression, making it practical for soft mechanism and actuation design.

The resulting design is depicted in Figs. 1 and 2. The monolithic structure consists of several functional elements printed simultaneously and compactly without assembly. These elements are: a parallel origami waterbomb mechanism with flexible hinges for producing 4-DoF motion, an end-effector (called the tactor) for transmitting forces to the fingerpad, eight vacuum-powered actuators for folding joints, four fluidic channels for supplying vacuum, and a thimble interface for inserting the user's finger. The device is compact, 40 mm long and 20 mm wide, and has hollow structures in the thimble, channels, and actuators that make it lightweight at 13.7 g (without tubing). The following subsections describe each component in detail.

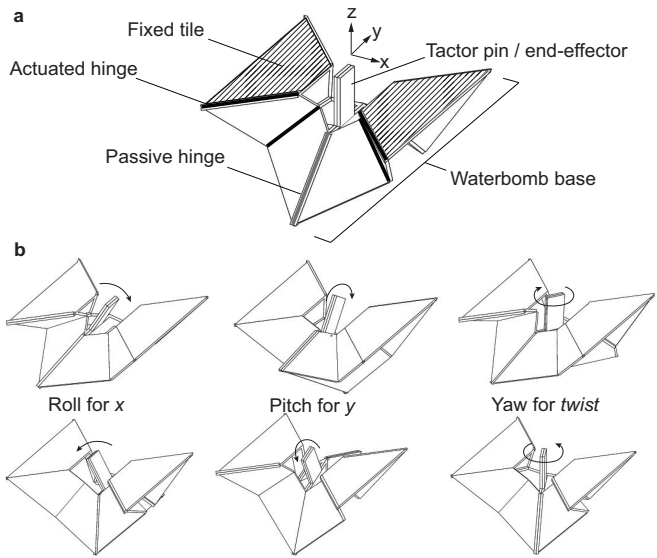

Fig. 3. The rigid origami model of the waterbomb base mechanism. It comprises two pairs of waterbomb base origami patterns connected in parallel at the end effector pin. Each pattern is a closed chain of six triangular tiles connected with six foldable hinges. The bold black lines indicate the actuated hinges. The pin produces 3 rotational DoF for roll (*x* motion), pitch (*y* motion), and yaw (twist) motions upon actuation. We achieve the fourth DoF, translation in the *z*-axis, utilizing the material softness.

*B. Origami-Inspired Multi-DoF Mechanism Design*

The primary design goal for *FingerPrint* is to produce lateral and rotational shear, normal pressure, and vibrational feedback for compelling interaction with virtual environments. Additionally, it should be compact and lightweight for minimal encumbrance. Origami-based foldable mechanisms offer great freedom toward multi-functionality, miniaturization, and assembly. Origami flexure hinges made of a compliant sheet are compact and require less assembly effort than classical two kinematic pair rotary joints [18]. We design an origami-inspired 4-DoF double waterbomb base parallel mechanism. The mechanism consists of two prefolded waterbomb base patterns connected in parallel (Fig. 3) for two-way motion and stability. Each pattern consists of six faces/tiles connected in a chain via eight bending hinges. The central tile connects both patterns, moves freely, and serves as the end-effector. Folding at any of the hinges causes roll, pitch, yaw rotational motion. The bold black lines indicate the actuated hinges. Several kinematic studies of waterbomb base mechanisms have been previously reported in the literature [19], [20]. We employ the mechanism's continuous roll and pitch rotational DoF for skin stretch in the lateral directions (*x* and *y* Cartesian coordinates) and its yaw motion for stimulating skin rotational shear. Although the rigid origami model here produces only three rotational DoFs, we achieve the fourth DoF in the *z*-axis for skin pressure by utilizing the material softness. Vacuum pressure applied to all actuators compresses the entire structure, so the tactor presses against the fingerpad. Rapid hinge folding and unfolding create vibrotactile haptic feedback.

*C. Distributed Actuation Design*

Actuating a mesoscale, multi-hinge origami mechanism is challenging in that it requires producing desired motions and forces with low encumbrance. Conventional electric motors are convenient but difficult to downsize. Thermally activated shape memory actuators (SMA) are powerful and compact but slow [21]. Dielectric elastomer-type actuators (DEA) are fast but require a large area to generate high forces [22]. Magnetic actuators can minimize encumbrance, but they necessitate large electromagnetic coils located in close proximity [23]. Piezoelectric actuators can achieve high frequencies but with small strokes [24]. To overcome this trade-off in force, speed, size, and stroke we developed a vacuum-powered foldable actuator that enables distribution and embedding on the waterbomb base structure as in Fig. 2. Unlike other most actuation technology, pneumatic actuation can be efficiently transmitted at a distance, thus the vacuum source can be located in the environment or worn on another location on the body. Additionally, pneumatic actuators are compatible with Magnetic Resonance Imaging (MRI) systems, making them effective for studying the neurological basis of tactile perception.

We propose a foldable actuator design in Fig. 2 that enables active high angular folding and passive unfolding due to the material elasticity. The applied vacuum pressure inside the actuator chamber collapses and folds it inward along the thin-walled hinges, which in turn produces a moment. The moment is a function of vacuum pressure that generates pulling forces on the surface area of the actuator inner walls [25]. The actuator restores its open state at the absence of pressure. Although the complete flexible model unknown, we designed the actuator geometry and size and operated them in pairs to increase the device's workspace, forces, and torques. The details of the pneumatic control are provided in Section IV.

*D. Thimble Interface Design for Grounding*

Proper grounding of a fingertip haptic device is necessary to deliver significant force on the fingerpad without unintentional noticeable reaction forces on and around the nail. To enable an effortless interface with the user's fingertip without additional straps, we developed a thimble structure that surrounds and fixates on the distal phalanx. Its hollow structure gives the tactor access to the fingerpad while the surrounding material distributes other contacts over the nail and neighboring skin, hence reducing noticeable reaction forces. Additionally, owing to its material flexibility and open circular ring design on the finger insert side, the thimble can expand slightly and accommodate various finger sizes.

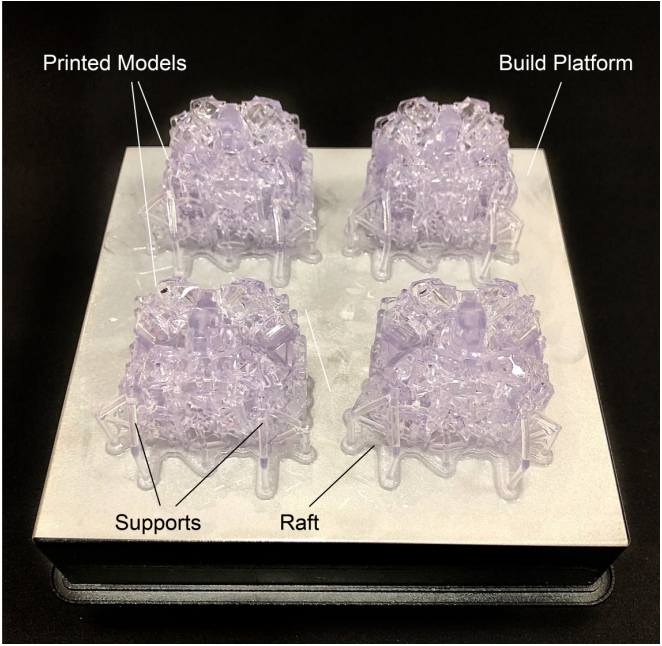

Fig. 4. Four copies of *FingerPrint* prototypes 3-D printed with scaffolding stems and raft supports in a single print process.

## III. Fabrication

We fabricated *FingerPrint* prototypes using the SLA 3-D printer and flexible resin described earlier. Here we explain the main steps, including model preparation and prototype post-processing, and cleaning procedures for achieving desired print quality.

### A. Model Preparation

The proposed design was assembled using the SolidWorks Computer-Aided Design (CAD) software from multiple individual parts and printed as a monolithic structure on Formlabs 3 3-D printer. The assembly CAD model (.SLDASM file) is first converted to a part file (.SLDPRT file) and then to a printer recognizable file format (.STL file). Then the file is imported to the printer's PreForm software program. As the printer cures resin layer-by-layer on the build platform upside down, the model is exposed to forces such as gravity and resin viscosity in the tank. It necessitates structural supports, namely rafts and scaffolding stems, to keep the model intact and preserve its structural integrity. We pay special care to the scaffold density, touchpoint size, and location, especially with our flexible models with fine features (Fig. 2). We minimize the support touchpoints on critical areas such as the thin actuator chamber walls and hinges to avoid added stiffness or damage after their removal. The model is supported around its perimeter and at the suspended regions, like the tactor and actuator external walls. We bypass supports inside the actuators and channels as they inhibit the actuator folding and airflow, respectively. In addition, we orient the model with the thimble facing toward the built platform to ensure its firm attachment through increased supports and to expose the waterbomb mechanism and resin bleed ports on the opposite side for the washing and cleaning processes described next.

### B. Post-Processing

Resin-based SLA prints involves washing, curing, and support removal post-processes. Depending on the model complexity, it also requires draining some of the resin trapped inside enclosed voids by providing access for washing liquid or removing residue with a syringe through dedicated holes. After printing, we place the build platform with parts still attached into a wash station (Formlabs Form Wash) with isopropyl alcohol for 15 minutes. Then we dry the samples and remove them from the platform. We suck the residual resin inside the actuators and channels by inserting a needle-tipped syringe barrel to the dedicated bleed ports and activating vacuum. Then the actuators and channels are flushed locally by pumping isopropanol (98%) into the bleed ports with a syringe with plunger. We alternate the wash and drain procedure 1-2 times until the resin is emptied. The transparent nature of the resin provides additional convenience for visually identifying the residue and its softness allows for navigating the needle to the desired areas.

The curing process involves exposing the model to UV light for a prescribed duration by adding stiffness and durability to the structure. Uniform light exposure is therefore crucial for the device's performance. Many supports in our printed prototype surround the thimble section that inhibits UV light access. We gently cut the supports off at the touchpoints using a cutter tool and Exacto knife. After, to seal the actuators, we plug all bleed ports by applying a drop of a flexible uncured resin. Four 1 mm diameter plastic tubes are inserted into the air supply ports and a liquid resin is applied at the interface for sealing. The prototype is then cured under UV light (405 nm wavelength) at 60°C heat for 6 minutes in the cure station (Formlabs Form Cure), which slightly stiffens the printed structure and solidifies the applied resin. The entire process takes about 30 minutes.

## IV. Experimental Characterization

To study and characterize the *FingerPrint*'s performance, we conducted three experiments. We measured the tactor's free range of motion, force and torque capacity in a blocked state, and frequency response for all four DoF. For these tests, we designed two experimental setups utilizing a stereo camera and a force sensor as described in the following subsection.

### A. Experimental setups

The experimental setup for measuring the tactor position, rotation, and frequency response is depicted in Fig. 5(a). It consists of a *FingerPrint* prototype mounted onto a 3-D printed rigid frame and a stereo camera positioned above the mechanism and oriented downward with the field of view orthogonal to the tactor. Owing to the constraints of directly tracking the tactor surface that is confined inside the thimble structure, we turned the prototype upside down and attached a marker on the bottom edge of the end-effector tile (bottom of the tactor) to track its free 3-D motion. The middle section of the end-effector tile is attached to the four hinges, so its protruding edge on the outside and extending tactor on the inside have similar length and pivot symmetrically opposite to each other. The four sets of eight actuators on the mechanism are driven independently by four miniature 3/2 solenoid valves as in Fig. 5(c) (LHDA0531115H, The Lee Company). The valve switching is controlled by an Arduino Uno microcontroller. We employ four thin (1 mm ID and 2 mm OD) and long (43 cm) silicone tubes to supply vacuum to the device. This ensures flexibility and reduced weight without compromising the airflow rate.

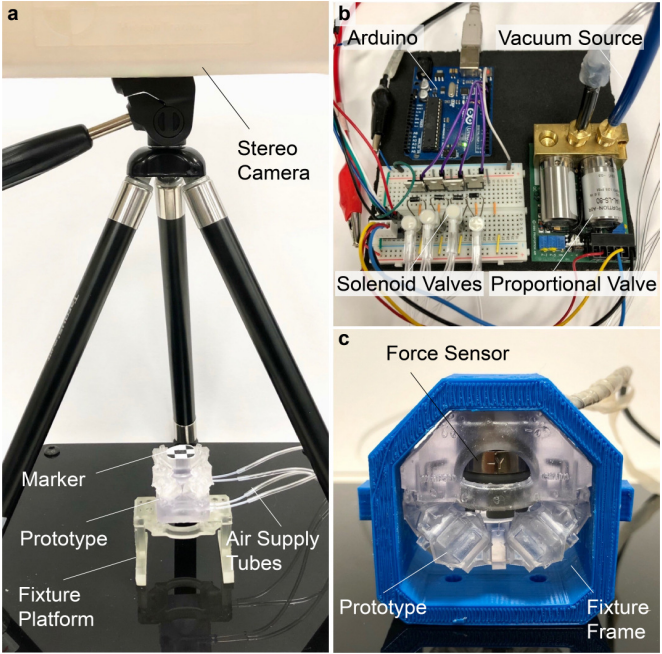

Fig. 5. Experimental setup and pneumatic control system employed for characterizing the device. (a) The free motion and frequency response setup by tracking with stereo camera. (b) The force/torque measurement setup with Nano 17 force sensor. (c) The pneumatic control system, which includes four miniature solenoid valves, a proportional valve and an Arduino board.

For these tests, we selectively activated the actuators on the device to measure two lateral tactile motions in the $x$ and $y$-axes that result from tactor roll and pitch rotations, linear motion in the $z$-axis and one twist motion (yaw) about the $z$-axis. We supplied square-wave pressure input with period of 2-3 seconds and amplitude of -90 KPa. Each motion was repeated three times and recorded on a stereo camera. To measure the device's frequency response, we applied square-wave pressure with frequency sweep from 1 Hz to 64 Hz and amplitude of -90 KPa.

The second experimental setup for measuring the mechanism's blocked forces and torques under varying input pressure is presented in Fig. 5(b). We attached a Nano-17 force sensor (ATI Industrial Automation) to the tactor via a 3-D printed rigid adapter. The adapter tightly fits onto the tactor and transmits forces to the force sensor. While we utilized the solenoid valves, we also employed a proportional valve (MM1/MM2, Proportion-Air, Inc.) to vary the vaccuum pressure and measure the tactor forces and torques. We applied periodic square-wave input pressure with amplitude from -27 KPa to -90 KPa. Each experiment was repeated three times.

### B. Free Motion Experiment Results

Snapshots of the tactor movement for 4-DoF are depicted in Fig. 6, and the tracking results for its range of motion are presented in Fig. 7. In Fig. 7(a), we plotted the $x$-$y$ Cartesian motions resulting from the rotational movement of the tactor in the corresponding axis ($x$ position for rotation around $y$-axis and $y$ position for rotation around $x$-axis). The $x$-displacement was approximately $\pm 3$ mm and the $y$-displacement was approximately $\pm 2$ mm. The $x$ displacement is larger due to the asymmetry of the waterbomb base mechanism. There is also a considerable hysteresis in the $x$ (11%) and $y$ (19.4 %) axes that we attribute to the undesired material deformations, mechanical imperfections, and non-uniform folding sequence of antagonistic actuators. The hysteresis could affect the ability to discriminate the direction of tactile stimuli. In future work we will reduce undesired structural deformation of tiles by further increasing their thickness and improve the overall backdrivability by designing bidirectional actuators. The tactor achieved a larger displacement of approximately 4 mm in the $z$-axis, as all eight actuators activate to push the tactor upward (see Fig. 7(b)). For twisting motion around the $z$-axis (yaw), the range was nearly $\pm 20$ degrees (see Fig. 7(c)). The tactor twist motion is coupled with $z$ linear motion, so it produces twisting torque and pressure force. We plan to improve the mechanism by decoupling the twist and linear motions and study the multi-axial tactor movement in arbitrary direction.

### C. Frequency Response

Fig. 8 displays the bandwidth of the tactor movement for 4 DoFs for square wave vacuum pressure input with sweeping frequency from 1 to 64 Hz. We calculate the magnitude in dB as $20\log_{10}\frac{d}{d_{max}}$, where $d$ is the amplitude of linear or angular displacement. The signal power reduces by half (-3 dB) above 10 Hz, whereas the bandwidth is lower for the twist motion around the $z$-axis. The translational DoF of the device demonstrate effectiveness for producing skin stimulation above the human volitional movement bandwidth ($\sim 10$ Hz). The device's frequency response is greatly affected by the flexible material viscoelasticity, the stiffness of the non-activated antagonistic actuators, and the switching speed of the pneumatic valve system. The frequency response range is within that of slowly adapting mechanoreceptors, such as Ruffini endings and Merkel's disk (<15 Hz), for sensing skin stretch and pressure, respectively. The tactor can reach high-frequency vibrations up to 64 Hz with low amplitudes, which are still detectable by the human skin with a reported sensing amplitude resolution of $\pm 50$ $\mu$m [26]. We believe the device's vibrational frequency can also activate rapidly adapting mechanoreceptors, such as Meissner's corpuscles ($<$50 Hz) that are sensitive to light touch and Pacinian corpuscles (optimal at 240 Hz).

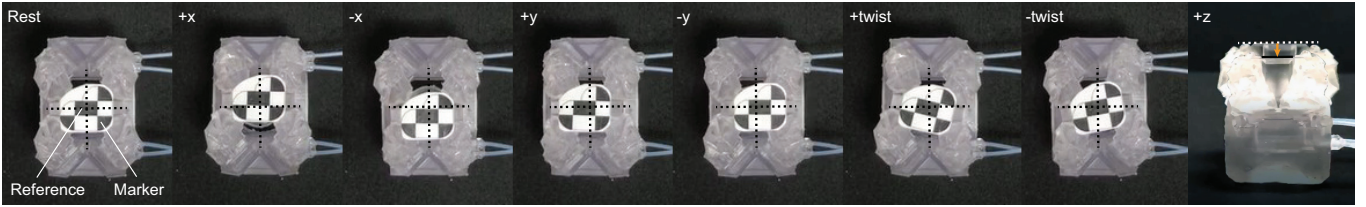

Fig. 6. Snapshots of the tactor movement in 4 DoFs. We employed a stereo camera to track the tactor motion in 3-D space using a marker. The virtual reference frame here aids to distinguish the marker's relative displacement.

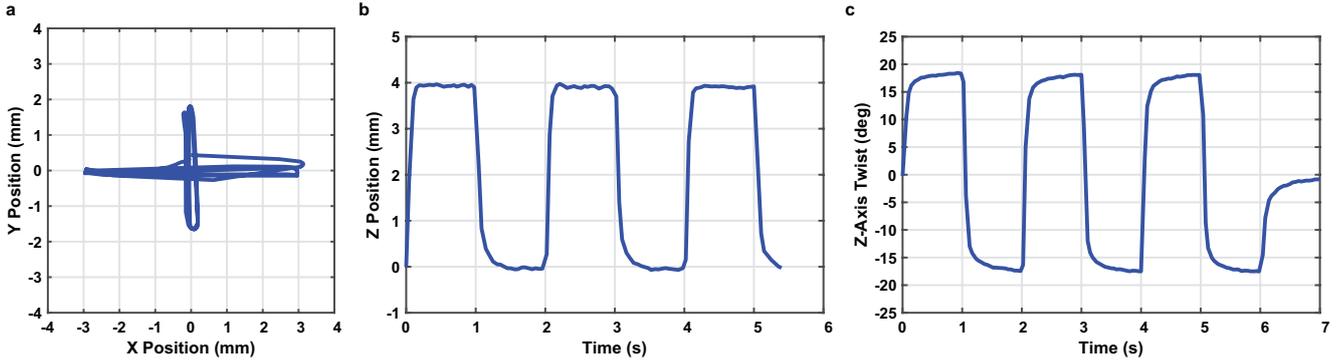

Fig. 7. Tactor free-motion tracking results for four DoF repeated three times. (a) The *x-y* position as a result of roll and pitch tactor motion, respectively. (b) The *z* position vs. time with tactor active displacement (1 s) and passive recovery (2 s). (c) The *z*-axis twist (yaw) motion vs.time.

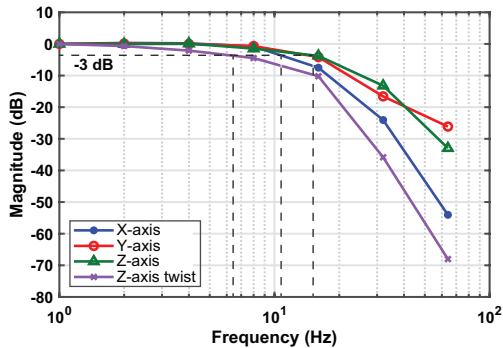

Fig. 8. The position bandwidth of the device for square wave vaccuum pressure input with varying frequency from 1 Hz to 64 Hz. The bandwidth for translational DoFs is above the approximate human volitional movement frequency of 10 Hz, whereas the twisting motion is lower.

*D. Blocked Force Experiment Results*

The experimental results for the blocked force and torque test are presented in Fig. 9. We observed similar force magnitude for the *x* and *y* axes (Fig. 9(a)), greater than $\pm 1$ N for maximum applied vacuum. As expected, the force output reduces with decreasing vacuum, demonstrating a tunable force output. The maximum *z*-axis force was higher, approximately 7 N (see Fig. 9(b). The twisting torque was nearly 20 N-mm in both directions (see Fig. 9(c). It is worth mentioning that the proportional valve controller causes small pressure oscillations that could result in noticeable haptic vibrations. We plan to employ valves with faster response in the future.

## V. CONCLUSION AND FUTURE WORK

We presented a novel, fully 3-D printable fingertip haptic device that produces 4-DoF motions and forces for mesoscale cutaneous tactile stimulation. The proposed method enables rapid manufacturing of miniature yet complex mechanisms with minimal assembly. While we demonstrated the effectiveness of the method for building miniature haptic interfaces with adequate performance, there exist several design limitations that we plan to address in the future. To achieve more compelling haptic stimulation, we plan to kinematically decouple twist and *z* movements. Furthermore, the print material should be further studied to improve the device's bandwidth for vibrotactile feedback. We plan to study multi-finger interaction and grasp with multiple devices in virtual environments, for example, on the index and middle fingers, and the thumb. We aim at further miniaturizing the device to minimize the physical interference between devices. We also intend to relocate the solenoid valves and electronics to the user's body to minimize the amount of tubing.

## ACKNOWLEDGMENTS

The authors thank Josue Gil-Silva for assisting with the fabrication process of early prototypes.

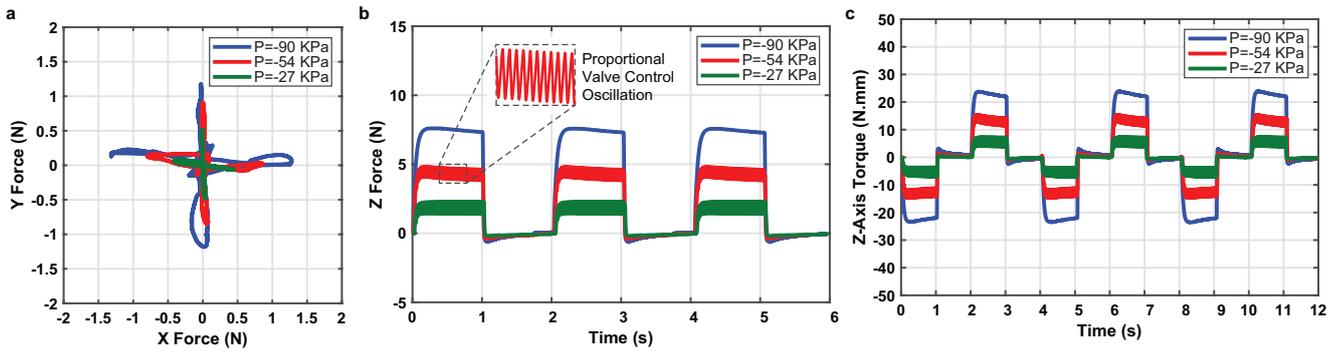

Fig. 9. The tactor's blocked force/torque measurement results for varying square-wave pressure input of -27 KPa, -54 KPa, and -90 KPa, each repeated three times. (a) The *x-y* force measurement plot. (b) The z force plot. (c) The *z*-axis torque. All DoF show increase in the output force with the increase in pressure.